\def\year{2020}\relax
\documentclass[letterpaper]{article} 
\usepackage{aaai20}  
\usepackage{times}  
\usepackage{helvet} 
\usepackage{courier}  
\usepackage[hyphens]{url}  
\usepackage{graphicx} 
\usepackage{graphicx}  
\usepackage{amsmath}
\title{An Auxiliary Classifier Generative Adversarial Framework for  Relation Extraction }

\begin{document}

\author{Yun Zhao \\
  Department of Computer Science \\
  University of California, Santa Barbara \\
  Santa Barbara, CA 93106 USA \\
  {\tt yunzhao@cs.ucsb.edu} \\}

\maketitle

\begin{abstract}
Relation extraction models suffer from limited qualified training data. Using human annotators to label sentences is too expensive and does not scale well especially when dealing with large datasets. In this paper, we use Auxiliary Classifier Generative Adversarial Networks (AC-GANs) to generate high-quality relational sentences and to improve the performance of relation classifier in end-to-end models. In AC-GAN, the discriminator gives not only a probability distribution over the real source, but also a probability distribution over the relation labels. This helps to generate meaningful relational sentences. Experimental results show that our proposed data augmentation method significantly improves the performance of relation extraction compared to state-of-the-art methods.
\end{abstract}

\section{Introduction}
Relation extraction aims to predict attributes and relations for entities in a sentence, which plays an essential role in information extraction~\cite{Survey_Relation}. Relation extraction models like Automatic Content Extraction (ACE)~\cite{ACE} use supervised learning methods, which suffer from limited high-quality training data. Traditional supervised approaches utilize human-labeled data, the quantity of which is far from being enough. Labeling sentences using human label is too expensive and not scalable especially when confronted with large datasets like NYT-Freebase~\cite{NYT}. Alternative paradigms are weakly-supervised learning methods such as distant supervision~\cite{Distant_supervision}. However, distant supervision is noisy, which could result in incorrect labeling. There are also some recent studies on Convolutional Neural Network (CNN) models for relation extraction~\cite{Zeng1,Zeng2}, among which Piecewise Convolutional Neural Network (PCNN) based methods~\cite{Zeng2} represent the state of the art.

Generative Adversarial Networks (GANs)~\cite{GAN} and their variants such as CGAN~\cite{CGAN}, InfoGAN~\cite{Infogan} and AC-GAN~\cite{AC-GAN} are appearing as very promising techniques for data generation, especially in the Computer Vision domain~\cite{DCGAN}. Text generation has not achieved equal success due to its discrete property in words. However, recent studies have proposed several approaches to deal with this problem. A reinforcement learning (RL) algorithm is used to upgrade the generator with reward signals instead of backpropagation in~\cite{Seqgan,Adversarial_gen}. Continuous samples are generated by manipulating the temperature of the softmax function and annealling to discrete values via the training process in~\cite{G_sequences,Categorical}.

In the AC-GAN architecture, class label information was added to both the generator and the discriminator~\cite{AC-GAN}. The objective function also has two parts: the cost of the real source and the cost of the correct class. Using AC-GAN as an unsupervised learning approach could solve the lack of training data problem by accurately generating relational sentences. The key contributions of this paper include: 

1) We propose an end-to-end AC-GAN based relation extraction framework, providing more high-quality training data for relation classifiers. 

2) Our generator network outputs the positions of the subject and object entities of the Resource Description Frameworks (RDFs), in addition to the next token every time step.

3) Experiment results show that our proposed data augmentation method improves the area under the curve (AUC) of the precision recall (PR) curve by 7.66\% compared to a PCNN, on NYT-Freebase.

\section{AC-GAN for Relation Extraction}

\begin{figure}[ht]
  \includegraphics[width=\linewidth]{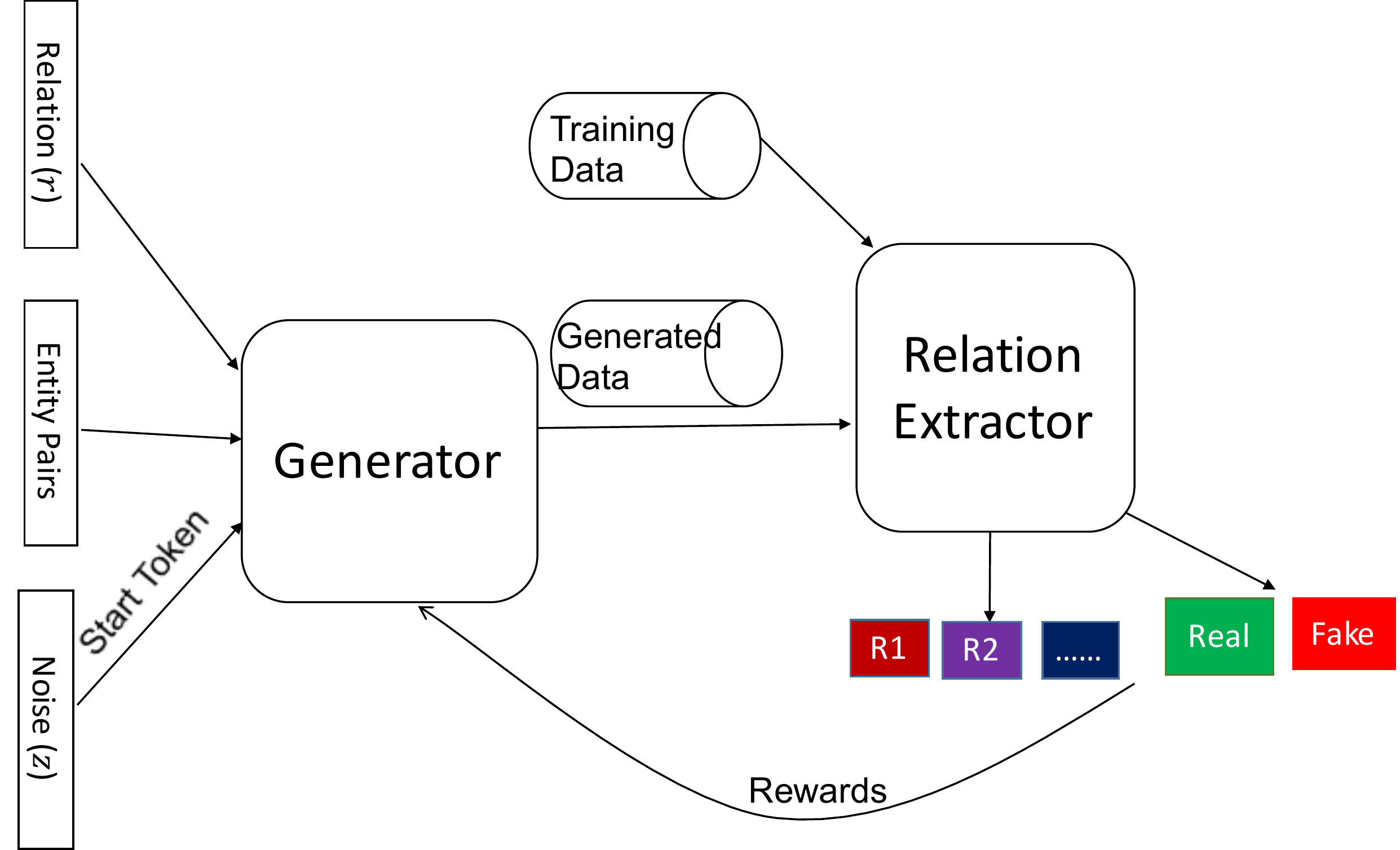}
  \caption{AC-GAN framework for relation extraction.}
  \label{GAN_model}
\end{figure}

Our model consists of a generator and a discriminator, shown in Fig.~\ref{GAN_model}. The generator is a Long Short Term Memory (LSTM) network with an embedding layer. The discriminator is a CNN based relation extractor, which outputs both a probability distribution over the real source and a probability distribution over the relation labels. To deal with the discrete value issue for GAN in Natural Language Processing (NLP), our sequence generation process is modeled as a sequential decision making process similar to SeqGAN~\cite{Seqgan}. The generator is treated as the agent of RL: The state is the generated tokens so far. The action is the next token to be generated. The reward is the estimated probability of the generated relational sentence being real and good to guide the generator. Note that the discriminator can only estimate reward of a finished sequence, so Monte Carlo Search is employed to obtain the average reward for current state.

\subsection{Generator}

\begin{figure}[h]
  \includegraphics[width=\linewidth]{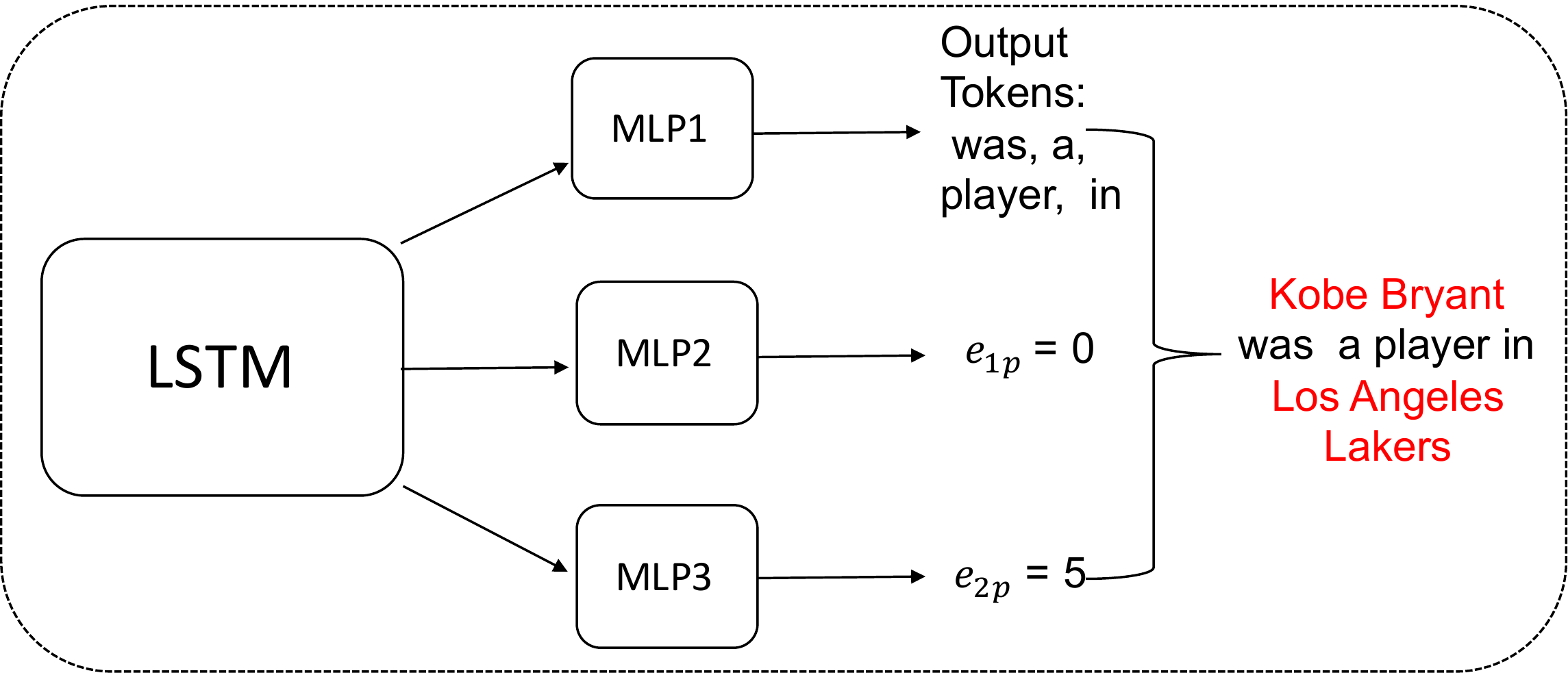}
  \caption{Generator model.}
  \label{Gen_model}
\end{figure}

The generator is a Recurrent Neural Network with three output layers, as is shown in Fig.~\ref{Gen_model}. In addition to the next token generated by decoders every time step, we introduce two more multilayer perceptron (MLP) layers which regress to the positions of the subject and object entities of the RDFs. For example, the generator generates the sentence of 'Kobe Bryant was a player in Los Angeles Lakers', which combines the information from the three outputs: output tokens, entity1 position ($e_{1p}$) and entity2 position ($e_{2p}$). The network structure is inspired by the Faster R-CNN method of object detection in Computer Vision~\cite{Faster_R_cnn}. The generator exhibits multi-task learning by hard parameter sharing among the three output branches. The losses from the three branches are added together and optimized. 

\subsection{Discriminator}
We build the discriminator on existing relation extractor, PCNN~\cite{Zeng2}. Since the generator needs rewards from discriminator to update, the discriminator gives both a probability distribution over the real source and a probability distribution over the relation labels. The loss function $L_D$ consists of the log-likelihood of the real source, $L_S$, and the log-likelihood of the correct relation label, $L_R$.
\begin{equation}
L_D=L_S+L_R.
\label{Loss_D}
\end{equation}

\begin{equation}
L_S=E[\log P(S=real|X_{real})]+
E[\log P(S=fake|X_{fake})].
  \label{Loss_D_S}
\end{equation}

\begin{equation}
L_R=E[\log P(R=r|X_{real})]+
	E[\log P(R=r|X_{fake})].
  \label{Loss_D_R}
\end{equation}

\subsection{Pretraining}
We give a warm start to the optimization by pretraining both the generator and the discriminator, since deep RL training is difficult to converge. The generator is pretrained with cross-entropy loss using the RDFs and sentences from the training dataset. The discriminator is pretrained using both real training data and sentences generated by the pretrained generator.

\subsection{Adversarial Training}
In our AC-GAN, every generated sentence has a relation label $r\sim p_r$, in addition to the noise $z$. GANs deal with a min-max game between generator and discriminator:
\begin{multline}
\min_{G}\max_{D}V(D,G)=E_{x\sim p_{data(x)}}[\log D(x)]+\\
E_{z\sim p_z(z)}[\log(1-D(G(z)))].
  \label{Loss_GAN}
\end{multline}
The generator tries to fool the discriminator by maximizing the expected reward gotten from the discriminator:
\begin{equation}
  J_{\theta}=E[R_{T}|s_0,\theta]=\sum_{t=1}^{T}\sum_{y_t\in y}G_\theta(y_t|s_{t-1})\cdot Q_{D_\phi}^{G_\theta}(s_{t-1},y_{t}).
  \label{Loss_Gen}
\end{equation}
According to REINFORCE algorithm~\cite{sutton1998reinforcement}, the gradient of $J_{\theta}$ is approximated using the likelihood ratios:

\begin{multline}
    \bigtriangledown_\theta J_{\theta} \approx  \sum_{t=1}^{T}\sum_{y_t\in y}\bigtriangledown_\theta G_\theta(y_t|s_{t-1})\cdot Q_{D_\phi}^{G_\theta}(s_{t-1},y_{t})
  \\=  \sum_{t=1}^{T}\sum_{y_t\in y} G_\theta(y_t|s_{t-1}) \bigtriangledown_\theta \log G_\theta(y_t|s_{t-1}) \cdot Q_{D_\phi}^{G_\theta}(s_{t-1},y_{t})\\
  \\=  \sum_{t=1}^{T}E_{y_t\in G_\theta(s_{t-1},y_{t})}[\bigtriangledown_\theta \log G_\theta(y_t|s_{t-1}) \cdot Q_{D_\phi}^{G_\theta}(s_{t-1},y_{t})].
  \label{Gradient_Gen}
\end{multline}

The reward is generated by multiplying the probability of the corresponding relation and being real.

\begin{equation}
Q_{D_\phi}^{G_\theta}(s_{t-1},y_{t})=R_{relation}\cdot R_{real},
  \label{Reward}
\end{equation}
where $R_{relation}$ denotes the probability for corresponding relation and $R_{real}$ is the probability of the generated sentences being regarded as real by the discriminator.

\section{Experiments}
\subsection{Data}
We evaluate our framework on a widely used dataset that was developed by~\cite{NYT}. As is shown in Table~\ref{Dataset_table}, the dataset consists of 52 unique relations along with an 'NA' relation. The training dataset contains 570088 sentences, while the testing data has 172448 sentences. We filter out the extra long sentences in addition to the 'NA' labeled sentences to train the generator network. When generating samples we limit the generator to sample sentences for 'non-NA' sentences. To show the effectiveness of the generator, we use Semeval-2010 task 8 dataset~\cite{Semeval}, which contains 8000 semantic relation sentences about 9 relations.

\begin{table}[t]
  \caption{NYT-Freebase dataset statistics}
  \label{Dataset_table}
  \centering
  \begin{tabular}{ll}
  \hline

    \# Relations (including NA) & 53    \\
    \# Training sentences      & 570088     \\
    \# Testing sentences     & 172448    \\
    \hline

  \end{tabular}
\end{table}

\subsection{Implementation Details}

The word embeddings are from~\cite{lin2016neural}, with embedding size of 50. The hidden dimension of the LSTM generator is 120. Both the discriminator and the generator are trained with Adam optimizer~\cite{Adam} but slightly different learning rates($lr_{dis} = 10^{-4}$ and $lr_{gen} = 10^{-3}$). When generating rewards for the generated tokens, we use Monte Carlo Search and take the average value of the rewards over rolling out for six times. We implement the deep learning models using Tensorflow~\cite{tensorflow}. More hyperparameter settings are described in Table~\ref{Dataset}.

\begin{table}[t]
  \caption{Hyperparameter settings}
  \label{Dataset}
  \centering
  \begin{tabular}{ll}
    \hline
    Generator &  \\
    \hline
     Batch size      & 64     \\
     Adam learning rate     & $10^{-3}$     \\
     Relation embedding size     & 50    \\
     LSTM hidden dimension     & 120    \\
     Sequence length & 100 \\
     Scheduled sampling threshold & 0.5 \\
     Gradient clip threshold & 5.0 \\
    \hline
    Discriminator &  \\
    \hline
     Batch size      & 64     \\
     Adam learning rate     & $10^{-4}$    \\
     Word embedding size     & 50    \\
     Position embedding size     & 5    \\
    \# Filters      & 128     \\
     Filter window size     & 3    \\
     Dropout keep probability & 0.5 \\
    \hline
    Rollout &  \\
    \hline
    Rollout number & 6    \\
    \hline
  \end{tabular}
\end{table}

During adversarial training, we use teacher forcing similar to~\cite{Categorical}. When the generator receives low rewards at the beginning of adversarial training, it is insufficient to update the generator purely based on the rewards. The generator needs to see more real training data. In teacher forcing, we apply maximum likelihood estimation (MLE) principle to update the generator more smoothly.

The training process consists of the following parts:

\begin{itemize}

\item Pretrain the generator using MLE on training dataset.

\item Generate sentences using the pretrained generator.

\item Pretrain the discriminator using the generated data and the real training data.

\item Jointly train the generator and the discriminator.

\end{itemize}

\subsection{Held-out Evaluation on NYT-Freebase}
\begin{figure}[h]
  \includegraphics[width=\linewidth]{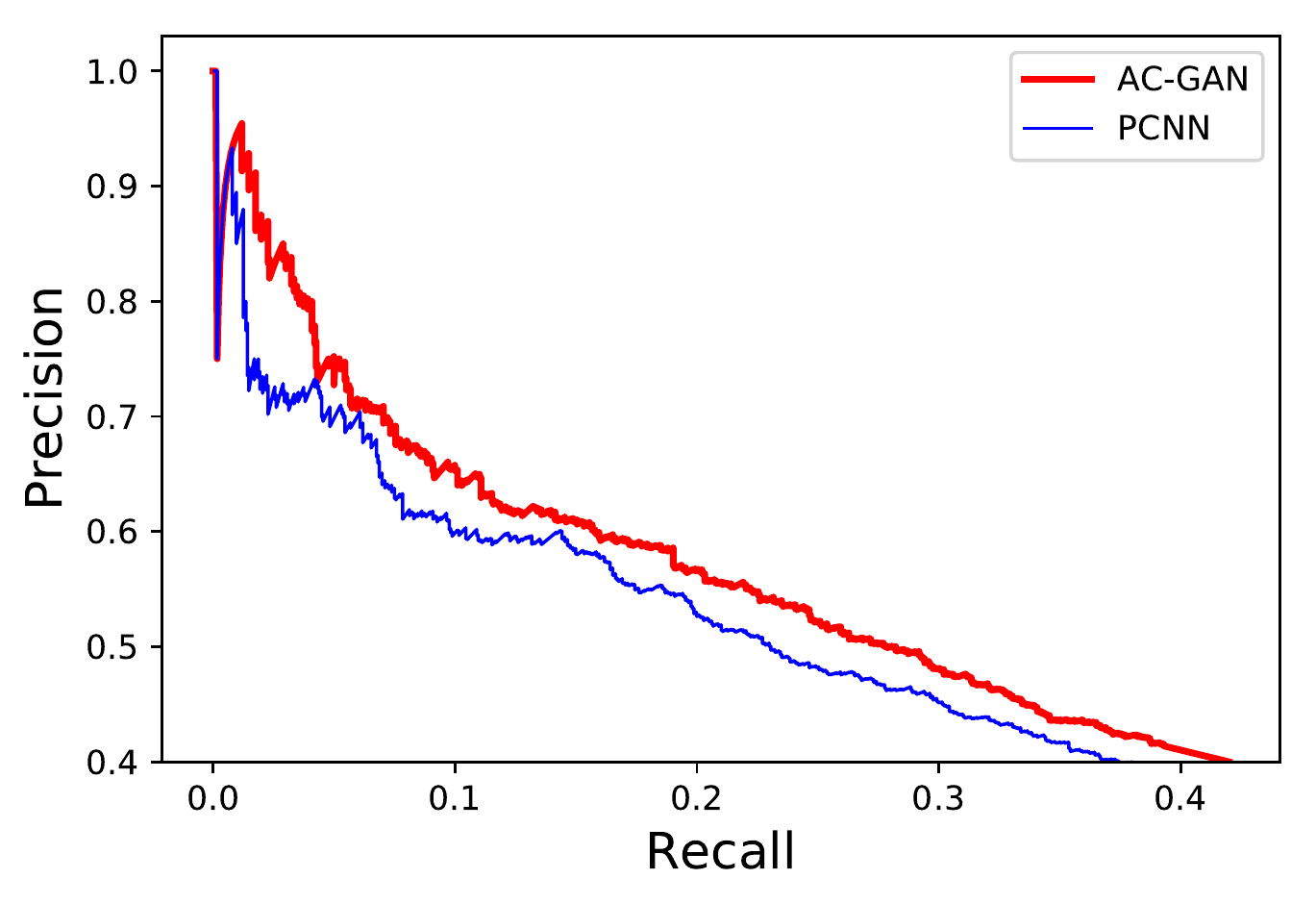}
  \caption{Performance comparison of proposed AC-GAN framework with PCNN.}
  \label{PR_curve}
\end{figure}

We compare our proposed AC-GAN framework with PCNN on NYT-Freebase dataset. Similar to previous works~\cite{Zeng2,lin2016neural}, we evaluate the AC-GAN framework using the held-out evaluation. In Fig.~\ref{PR_curve}, experiment results indicate that AC-GAN consistently outperforms PCNN. Our proposed data augmentation method improves the AUC of PR curve by 7.66\% compared to PCNN. This is mainly due to the fact that using AC-GAN, the generator can consistently provide high-quality relational sentences.

\subsection{Generated Samples from AC-GAN trained on Semeval-2010 Task 8 Dataset}

Below are some sample relational sentences generated through AC-GAN trained on Semeval-2010 task 8 dataset. Sentences generated with and without discriminator are presented for comparison. We present the Cause-Effect and Product-Producer relation samples in the following.\\

\textbf{Cause-Effect Results from AC-GAN:}
\begin{center}
The Peru earthquake triggered Avalanche.\\
The vascular dilatation was caused by the course.\\
Poverty is caused by administration.\\
The cysts are caused by every kind of environment.\\
The course on a calm day led to more persons with motivation to run gains from long running.\\
\end{center}

\textbf{Cause-Effect Results from LSTM only:}
\begin{center}
The high humidity caused by the bacteria damaged. \\
The output voltage swing obtainable from the anger is caused by commission from the mug and a huge problem in China.\\
The decline has always lowing, to take feather samples. \\
Many more resources are directly the fifth century were almost easy to identify and the late 1990s. \\
His highly original and topical act disorder encouraged the drama has fever from constipation whereas tricks and sent liable for the Jewish people. \\
\end{center}

\textbf{Product-Producer Results from AC-GAN:}
\begin{center}
This comment refers to this dissertation proposal.\\
Top problems can center attacks.\\
Here was the author of the book.\\
This comment refers to the almost invisible year of Ste90. \\
A company manufactures books in the story of 85. \\
The message was generated by the scientist. \\
\end{center}

\textbf{Product-Producer Results from LSTM only:}
\begin{center}
In 1890, the Banbury has constructed a valid plan beans, the human face, from in order being with their responsibilities.\\
The 22-year-old rapper posted a message on his stems that Christmas.\\
The authors who center the wrong book is 3000 well as a method of two call out of Christmas.\\
The 22-year-old rapper posted a hill.\\
Toda has dropped the formal announcement.\\
The man has dropped the townhouse.\\
\end{center}

Other relations (like Instrument-Agent etc.) have similar results. It's obvious that sentences generated from AC-GAN are more meaningful compared with sentences generated from LSTM, which could help in relation extraction.

\section{Conclusion}

In this paper, we propose an AC-GAN based framework to generate meaningful relational sentences for relation classifiers. Experiment results show that the AC-GAN framework significantly improves relation extraction performance.



\bibliographystyle{aaai}
\bibliography{main}
\end{document}